\documentclass[final,12pt]{elsarticle}

\usepackage{amssymb}
\usepackage{lineno}

\usepackage{graphicx}
\usepackage{amsmath}
\usepackage{amssymb}
\usepackage{hyperref}
\usepackage{algorithmic}
\usepackage{subfig}

\begin{document}

\begin{frontmatter}

\title{TraX: The visual Tracking eXchange Protocol and Library}
\author{Luka \v{C}ehovin}

\address{Faculty of Computer and Information Science, University of Ljubljana}

\begin{abstract}
In this paper we address the problem of developing on-line visual tracking algorithms. We present a specialized communication protocol that serves as a bridge between a tracker implementation and utilizing application. It decouples development of algorithms and application, encouraging re-usability. The primary use case is algorithm evaluation where the protocol facilitates more complex evaluation scenarios that are used nowadays thus pushing forward the field of visual tracking. We present a reference implementation of the protocol that makes it easy to use in several popular programming languages and discuss where the protocol is already used and some usage scenarios that we envision for the future.
\end{abstract}


\end{frontmatter}


\section{Introduction}
\label{}

Visual tracking is an active and diverse research area in computer vision which deals with the question how to predict a state of one or more objects in a sequence of images. This problem is difficult to tackle as a whole and has therefore sprouted numerous different approaches with different tracking scenarios, constraints, methodologies for performance evaluation, and techniques for algorithm analysis. Integrating tracking algorithms with evaluation tools requires at the very least some tedious source code adaptation. Moreover, classical batch evaluation approaches, that come from classification and detection methodologies are very limiting because of complex sequential nature of the tracking problem.

In this paper we present a technical solution to these problems. We have designed a specialized protocol called TraX, which stands for {\em visual Tracking eXchange}. The protocol (Section~\ref{sec:protocol}) defines a standardized way of integrating tracking algorithms with applications, without requiring access to the source code and therefore facilitates software re-use and consolidation. We present a reference cross-platform implementation of the protocol (Section~\ref{sec:library}) that simplifies its adoption, list its features and discuss existing and potential use cases (Section~\ref{sec:features}), provide an example of integration (Section~\ref{sec:example} and \ref{sec:application}) and conclude with a summary and development road-map (Section~\ref{sec:conclusion}).

\section{Problems and Background}
\label{}

A common challenge when evaluating a newly developed tracking algorithm is ensuring that the results and methodology are comparable with existing work. Results in papers are trimmed down due to paper length limitations, therefore the experiments often have to be repeated to ensure a meaningful comparison. This requires access to reliable implementations of the algorithms. Because of this, large-scale comparisons using a consistent methodology were in the past a challenging manual task of adapting or implementing trackers and typically resulted in standalone survey papers~\cite{Wang2011,Pang2013,Wu2013,Smeulders2013}. Recently, more focus has been directed towards designing consistent evaluation methodologies~\cite{Wu2010,Cehovin2016,Smeulders2013}. Wu et al.~\cite{Wu2013} presented a benchmark dataset and provided a toolkit for performance evaluation of detection-based trackers. To use this toolkit, one has to adapt their tracker as well as modify the toolkit's code. In~\cite{Smeulders2013} authors propose only the methodology, but leave its implementation to researchers, which makes its adoption difficult. Perhaps the most advanced in this respect is the VOT Challenge initiative~\cite{Kristan2016} that started in 2013 and organizes an annual challenge and a workshop based on the results of a challenge. From the technical point of view, the core of the challenge is a toolkit that can perform various experiments on a large number of tackers.

Another important research problem is determining which tracking algorithms are suitable for a specific tracking task, which requires testing algorithms under various conditions. A handful of such task-specific tools were proposed, such as the ODViS system~\cite{Jaynes2002} which is focused on tracking in a surveillance system scenarios, or the work by Nawaz and Cavallaro~\cite{Nawaz2012}, who proposed a system that simulates several sources of noisy input, such as initialization noise, per-pixel image noise, and changes in the frame-rate. Both systems require direct integration of tracking algorithms into the evaluation code-base which is time-consuming, limits the choice of programming language, and presents a software stability risk.

The issue of consistent evaluation and its repeatability is partially addressed if the authors share their implementation, either as source code or in a binary form. However, there is still a lot of work in making these implementations run with a specific methodology or application. In case of binary versions it can be even impossible to properly run the tracker beyond its predefined purpose. Evaluation methodologies are also becoming more ambitious, but also more complex and implementing them multiple times is error-prone. From the perspective of the research community, the best strategy is to encourage development of common open-source evaluation tools and focus on their stability and reliability. This is where our work steps in. The proposed protocol acts as a standardized bridge between the tracker and the utilizing application targeting primarily on-line tracking scenarios where consistent integration interface is especially difficult to make. Once conforming with the protocol, tracker implementations can be quickly used in various scenarios without additional modifications (or even access to the source code). The protocol is implemented in a multi-platform open-source library, which can be used to add the protocol support to a tracker implementation or to design a new use-case.

\section{The protocol}
\label{sec:protocol}

The TraX protocol is simple, yet flexible enough to allow extensions and custom use-cases. Still, all technical details of the protocol are too long for this presentation, and we refer the reader to~\cite{Cehovin2014a} and the on-line project documentation for more information.

\begin{figure}
    \centering
    \includegraphics[width = \linewidth]{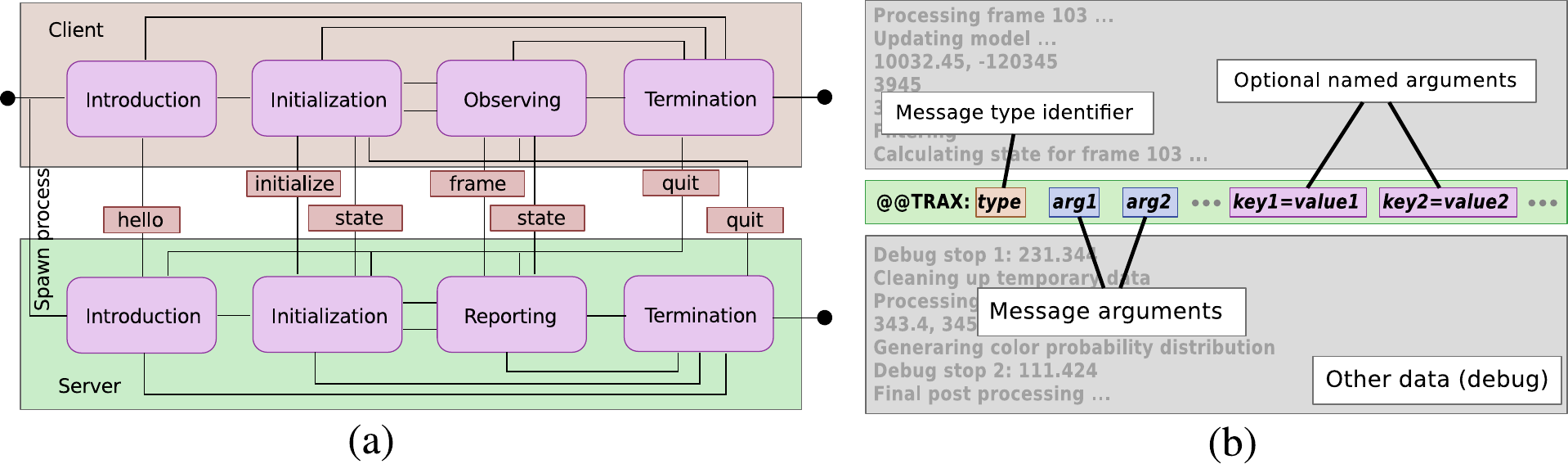}
    \caption{Protocol states for client and server (a) protocol message format (b).}
    \label{fig:protocol}
\end{figure}

To describe the interaction between the tracker and utilizing application, we adopt a standard client-server terminology. A {\em server} is the tracker process that answers to requests made by the client\footnote{Unlike traditional servers that communicate with multiple clients, the server in our case only communicates with a single client.}. The {\em client} is driving the tracking session; typically this would be an evaluation tool that aggregates tracking data for performance analysis. Client's requests may tell the tracker to initialize its model with initial object state in the given image, or to predict state of the object in a given image. The possible states of both parties and the messages that trigger change are visualized in ~\ref{fig:protocol}(a). The synchronous communication allows clients to select images based on the trackers output, opening new possibilities in tracker evaluation, which, up until now was limited by batch execution of tracker program.

The protocol is designed to use standard input and output streams, a mechanism that all modern operating systems provide, to exchange line-based messages between the tracker process and evaluation application, embedded between normal tracker output (e.g. debug messages). Other media, such as TCP streams can be used as well, enabling remote and distributed evaluation. Each protocol message is separated from the past and future stream content by the new-line character. A fixed message prefix is used to distinguish between arbitrary program outputs and embedded TraX messages. The prefix is followed by a message type identifier, which is followed by message arguments. A protocol message, illustrated in Figure~\ref{fig:protocol}(b), consists of a header and mandatory arguments as well as optional parameters that can carry additional data. The two core data-types conveyed by the protocol are image (filesystem path, in-memory data) and target state (e.g. bounding box, polygon).

\section{Reference library}
\label{sec:library}

\begin{figure}
    \centering
    \includegraphics[width = 0.6\linewidth]{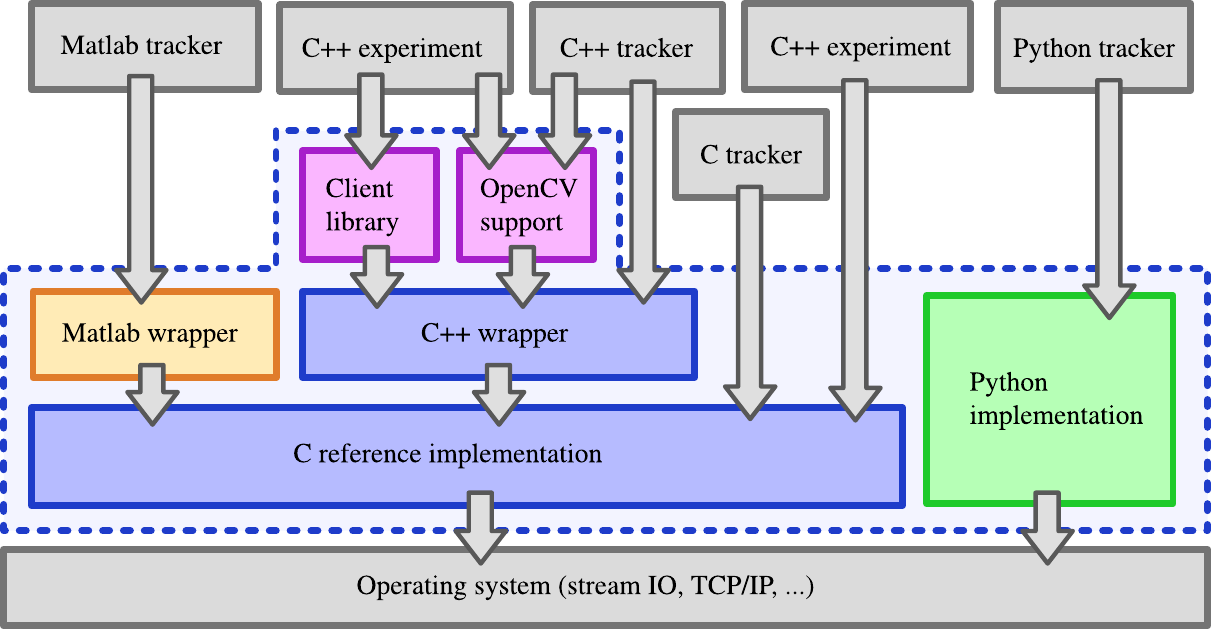}
    \caption{Overview of the library infrastructure.}
    \label{fig:architecture}
\end{figure}

The TraX protocol is simple to implement in its basic form, however, its full and consistent implementation requires more time. To make the adoption more convenient we have created a reference implementation library that hides the protocol details and can be used to implement support in servers (trackers) as well as in clients. The library is written in pure C\footnote{The reason for this is that C code is easy to bind into other languages.}, is cross-platform and does not require any external dependencies. We also provide C++ and Matlab/Octave wrappers that expose the underlying functionality in a more friendly manner, a Python implementation of the protocol, and utility libraries that handle some common use cases, e.g. integration with the OpenCV library or writing a client application. Figure~\ref{fig:architecture} illustrates the architecture of the entire project.

\subsection{Functionality and use cases}
\label{sec:features}

There are two ways to use the library: (1) include the support in your tracker implementation, as demonstrated in Section~\ref{sec:example}, (2) write a client, i.e., program that drives the tracking session. We envision several client applications, some of which go beyond the current state of tracker evaluation. The most typical client application performs comparative experiments in objective and repeatable manner. In fact the proposed protocol (together with the reference library) was recently adopted as the default means of integration in VOT Challenges, significantly shortening the time of experiment execution\footnote{By definition, supervised experiments require reinitialization on tracker failure, which can be detected immediately when using the proposed protocol.}. Because of its flexibility, the protocol may also be used to test a tracker under different combinations of parameters (passed to the tracker as extra parameters during initialization) or to aggregate and analyze tracker-specific information (passed to the application for each frame as extra parameters during tracking), helping researchers to understand its behavior.

Moving beyond classical performance benchmarking, the most important potential of the protocol lies in its synchronized exchange of information, which could be used to simulate real-time requirement for algorithms (e.g. dropping frames), as well as simulate scenarios of active tracking, where the response of the tracker influences the content of the subsequent images.

\subsection{Tracker integration}
\label{sec:example}

In Figure~\ref{fig:example} we show how a typical tracker can be extended to support the protocol. For brevity, we use high-level pseudo-language. More realistic and complete examples in various programming languages can be found in the project documentation.

\begin{figure}[h]
\begin{minipage}[t]{0.5\linewidth}
{\tiny
\begin{algorithmic}
\STATE Setup tracker
\STATE Read initial object region and first image
\STATE Initialize tracker with provided region and image
\LOOP
    \STATE Read next image
    \IF{ image is empty}
        \STATE Break the tracking loop
 	\ENDIF
    \STATE Update tracker with provided image
	\STATE Write region to file
\ENDLOOP
\STATE Cleanup tracker
\end{algorithmic}
}
\end{minipage}
\begin{minipage}[t]{0.5\linewidth}
{\tiny
\begin{algorithmic}
\STATE Setup tracker
\STATE {\em TraX:} Initialize protocol, send introduction
\LOOP
    \STATE {\em TraX:} Wait for message from client
	\IF{initialization request}
		\STATE Initialize tracker with provided region and image
	\ELSIF{frame request}
		\STATE Update tracker with provided image
	\ELSIF{terminate request}
		\STATE Break the tracking loop
	\ENDIF
    \STATE {\em TraX:} Report current state
\ENDLOOP
\STATE {\em TraX:} Cleanup protocol
\STATE Cleanup tracker
\end{algorithmic}
}
\end{minipage}
    \caption{Pseudo-code of a tracker loop before (left) and after (right) protocol integration.}
    \label{fig:example}
\end{figure}

The major modification of the original tracking loop in the example is that tracker initialization is moved within the tracking loop as the client may request reinitialization during the tracking session. The tracker only has to take care of serving the incoming requests which takes away a lot of boilerplate code, e.g., loading image sequence and ground-truth.

\subsection{Client applications}
\label{sec:application}

In Figure~\ref{fig:example_client} we show how a generic client application in pseudo-language. According to the TraX protocol the client application has to manage the tracker process, how this is done is not specified by the protocol and depends on the platform and the language used. A C++ support library for the reference protocol implementation that takes care of platform specifics. A detailed tutorial on how to use this library is available in the on-line documentation.

\begin{figure}[h]
{\tiny
\begin{algorithmic}
\STATE Start tracker process
\STATE {\em TraX:} Initialize protocol, wait for introduction
\LOOP
    \STATE Obtain sequence frame
	\IF{not initialized}
		\STATE Obtain ground-truth for the current frame
        \STATE {\em TraX:} Send initialize command with frame data and ground-truth
	\ELSE
		\STATE {\em TraX:} Send update command with frame data
	\ENDIF
	\IF{tracker terminated}
		\STATE Break the tracking loop
    \ELSE
		\STATE Inspect tracker's output and react
	\ENDIF
\ENDLOOP
\STATE {\em TraX:} Cleanup protocol
\STATE Cleanup client
\end{algorithmic}
}
    \caption{Pseudo-code of a simple client application according to the TraX protocol.}
    \label{fig:example_client}
\end{figure}

\section{Conclusions}
\label{sec:conclusion}

We have presented a specialized communication protocol that separates the development of on-line tracking algorithms from the development of tools that are used to drive the tracking session. This allows researchers to focus on the development of new algorithms without implementing the boilerplate code for evaluation or other uses. The presented version of the protocol and its corresponding implementation have some limitations, e.g. they can only be used for single-target tracking and the state of the target can only be specified with a simple region. We plan to address these constraints in future versions together with support for more programming languages. We would like to emphasize that the protocol is also already being used in practice, most prominently with the adoption for tracker integration in the VOT Challenge evaluation toolkit~\cite{Kristan2016}.

\section*{References}

\bibliographystyle{elsarticle-num}
\bibliography{cehovin2016trax}

\end{document}